\documentclass[letterpaper, 10 pt, journal, twoside]{IEEEtran}
\usepackage{amsmath}
\usepackage{amsfonts}
\usepackage{algorithmic}
\usepackage{algorithm}
\usepackage{array}
\usepackage[caption=false,font=normalsize,labelfont=sf,textfont=sf]{subfig}
\usepackage{textcomp}
\usepackage{stfloats}
\usepackage{url}
\usepackage{verbatim}
\usepackage{graphicx}
\usepackage{cite}
\usepackage{bbding}
\usepackage{siunitx}
\usepackage{array}
\usepackage{booktabs}
\usepackage{tabularray}
\usepackage{array}
\usepackage{multirow}
\usepackage{colortbl}
\usepackage{hhline}
\usepackage{vcell}
\usepackage{ragged2e}
\usepackage{tabularx}
\usepackage[normalem]{ulem}
\usepackage{color}
\usepackage{xcolor}
\newcommand{\red}[1]{\textcolor{black}{#1}}

\begin{document}

\title{A Survey on Soft Robot Adaptability: Implementations, Applications, and Prospects}
\author{Zixi Chen$^{1,*}$, Di Wu$^{2,*}$, Qinghua Guan$^{3}$, David Hardman$^{4}$, Federico Renda$^{5}$, Josie Hughes$^{3}$,\\ Thomas George Thuruthel$^{6}$, Cosimo Della Santina$^{7, 8}$, Barbara Mazzolai$^{9}$, Huichan Zhao$^{10}$ and Cesare Stefanini$^{1}$
\thanks{This work was supported by the European Union by the Next Generation EU project ECS00000017 ‘Ecosistema dell’Innovazione’ Tuscany Health Ecosystem (THE, PNRR, Spoke 9: Robotics and Automation for Health.) \emph{Corresponding authors: Di Wu.} This work has been submitted to the IEEE for possible publication. Copyright may be transferred without notice, after which this version may no longer be accessible.} 
\thanks{
$^{*}$Equal Contribution.
$^{1}$The BioRobotics Institute and the Department of Excellence in Robotics and AI, Scuola Superiore Sant’Anna, Pisa, Italy.
$^{2}$Mærsk Mc-Kinney Møller Instituttet, University of Southern Denmark, Odense, Denmark.
$^{3}$CREATE Lab, EPFL, Lausanne, Switzerland.
$^{4}$Bio-Inspired Robotics Lab, University of Cambridge, Cambridge, UK.
$^{5}$The Khalifa University Center for Autonomous Robotic Systems, Khalifa University, Abu Dhabi, United Arab Emirates.
$^{6}$Department of Computer Science, University College London, London, United Kingdom.
$^{7}$Department of Cognitive Robotics, Delft University of Technology, Delft, The Netherlands.
$^{8}$Institute of Robotics and Mechatronics, German Aerospace Center (DLR), Wessling, Germany.
$^{9}$Bioinspired Soft Robotics, Istituto Italiano di Tecnologia, Genoa, Italy.
$^{10}$Department of Mechanical Engineering and Beijing Key Lab of Precision/Ultra-Precision Manufacturing Equipment and Control, Tsinghua University, Beijing, China.
}

}


\maketitle
\begin{abstract}
Soft robots, compared to rigid robots, possess inherent advantages, including higher degrees of freedom, compliance, and enhanced safety, which have contributed to their increasing application across various fields. 
Among these benefits, adaptability is particularly noteworthy. 
\red{In this paper, adaptability in soft robots is categorized into external and internal adaptability. External adaptability refers to the robot’s ability to adjust, either passively or actively, to variations in environments, object properties, geometries, and task dynamics. Internal adaptability refers to the robot’s ability to cope with internal variations, such as manufacturing tolerances or material aging, and to generalize control strategies across different robots.}
As the field of soft robotics continues to evolve, the significance of adaptability has become increasingly pronounced. 
In this review, we summarize various approaches to enhancing the adaptability of soft robots, including design, sensing, and control strategies. 
Additionally, we assess the impact of adaptability on applications such as surgery, wearable devices, locomotion, and manipulation. 
We also discuss the limitations of soft robotics adaptability and prospective directions for future research. 
By analyzing adaptability through the lenses of implementation, application, and challenges, this paper aims to provide a comprehensive understanding of this essential characteristic in soft robotics and its implications for diverse applications.
\end{abstract}

\section{Introduction}
\label{sec1}

Soft robotics is a rapidly evolving field with an ever-expanding range of applications. 
Due to the inherent flexibility and compliance, adaptability is a significant advantage of soft robots. 
Early research in soft robot adaptability primarily explored the applicability of soft robots across various external environments and with different objects.
For example, the complex nature of human anatomy necessitates that soft robots can conform to intricate bodily lumens while maintaining safe interaction forces \cite{cianchetti2014soft}. 
Similarly, soft locomotion robots have been tested on diverse terrains, e.g., different ground hardness \cite{bosworth2016robot} and humidity \cite{yang2021starfish}.
In addition to hardware, algorithms also play a crucial role in maximizing the adaptability of soft robots.
Intelligent strategies such as reinforcement learning (RL)  and learning from demonstration (LfD) endow robots with the ability to manipulate various objects \cite{gupta2016learning} and achieve various tasks \cite{jiang2021hierarchical}.
Furthermore, instead of proposing dedicated sensing and control strategies for each hardware platform, it is crucial to propose adaptive algorithms that function across multiple platforms.
\red{Recently, researchers have begun developing algorithms that are robust to changes in robot hardware due to aging and transferable to hardware with different configurations.}
For instance, an adaptive controller can be utilized for robots with different stiffness caused by manufacturing tolerances \cite{chen2023hybrid}. 
An adaptive calibration approach applicable to a series of sensors has been introduced in \cite{kim2020adaptive}, facilitating the sensor's mass production and long-term usage.

\begin{figure*}[!ht]
\centering
\includegraphics[width=\linewidth]{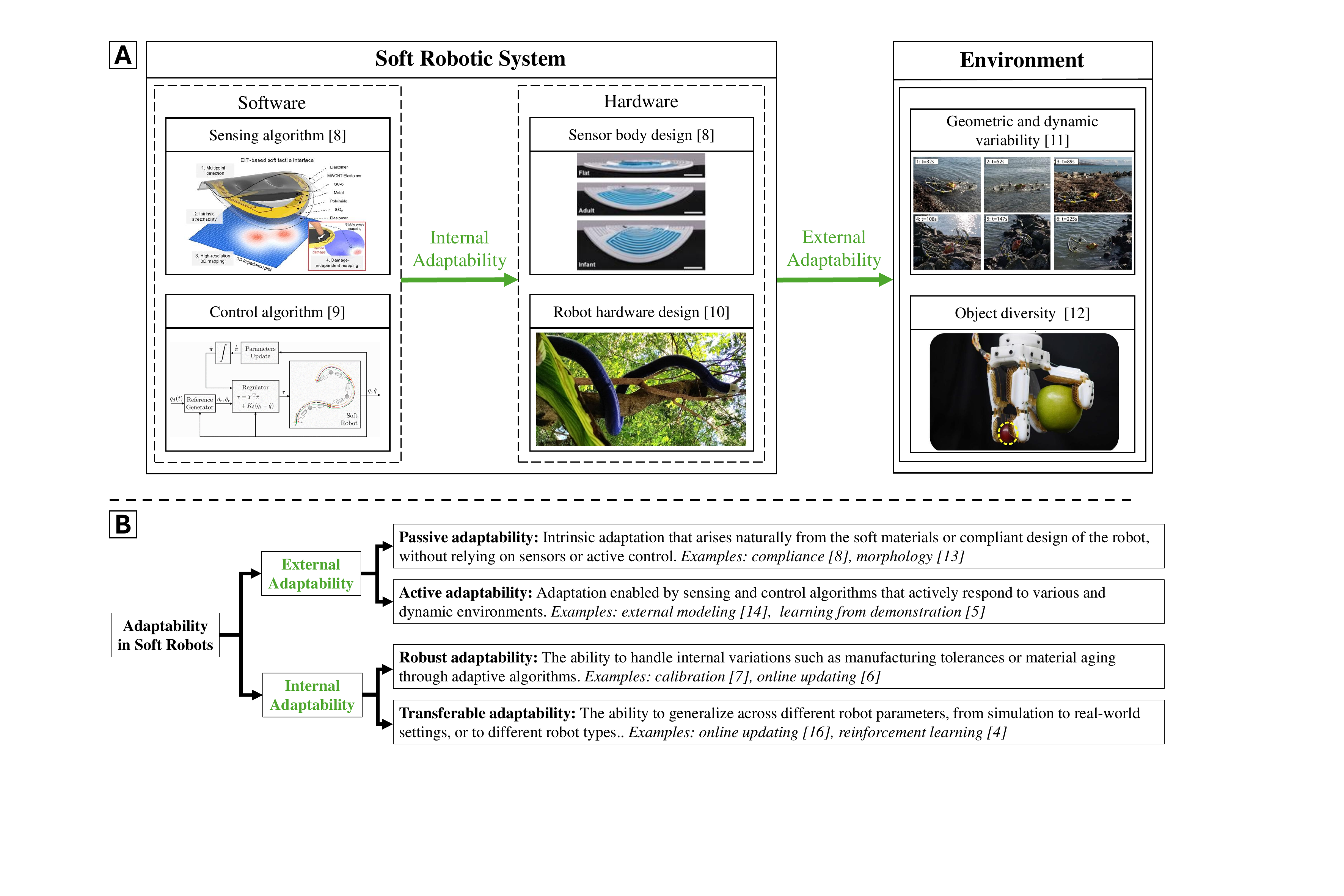}
\caption{
\red{(A) Adaptability in soft robotics can be categorized into external and internal adaptability.
External adaptability refers to a soft robot’s ability to adjust to dynamic and varied environments.
Internal adaptability refers to the ability of software to remain effective across soft robots with different parameters or configurations.
(From \cite{kim2024extremely, trumic2021adaptive, del2024growing, polzin2025robotic, ruotolo2021grasping}; used with permission)
(B) The classification of external and internal adaptability: 
Passive adaptability refers to the robot’s natural response to environmental changes and object interactions, primarily enabled by its inherent material properties or compliant structural design. 
In contrast, active adaptability involves deliberate, control-driven behaviors guided by intelligent algorithms to dynamically respond to external variations.
Internal adaptability encompasses the capacity of adaptive algorithms to maintain performance despite hardware variations resulting from material aging or manufacturing tolerances and to transfer effectively across robots with differing physical parameters or material compositions.}
}
\label{fig1}
\end{figure*}

Building on the aforementioned works that emphasize soft robot adaptability, we propose the following definition for soft robot adaptability:
\textit{Adaptability in soft robots refers to their ability to actively or passively adjust to variations and disparities in diverse and dynamic environments, objects, tasks, and robot components.}
\red{As illustrated in Fig. \ref{fig1}-(A), adaptability can be categorized based on its target: external adaptability, where soft robots adjust to different and changing objects and environments, and internal adaptability, where adaptive algorithms are shown to be robust to variations in hardware parameters caused by manufacturing tolerances or aging, and transferable to hardware with different configurations.}
Examples of external adaptability include soft locomotion robots that can navigate various external environments \cite{polzin2025robotic} and soft hands that can manipulate objects of different shapes \cite{ruotolo2021grasping}.
Internal adaptability is demonstrated in contact construction sensing algorithms showing robustness to soft sensors under various deformations \cite{kim2024extremely}, and adaptive control strategies dynamically update model parameters to transfer across soft robots with different stiffness levels \cite{trumic2021adaptive}.

\red{External adaptability can be further categorized into passive and active adaptability, as illustrated in Fig. \ref{fig1}-(B).
Passive adaptability refers to the soft robot’s intrinsic and natural response to environmental changes, primarily achieved through its soft materials or compliant structures. 
This is a characteristic unique to soft robots.
For example, soft grippers deform passively to conform to the shapes of different objects without dedicated active control \cite{zhou2017soft}.
In contrast, active adaptability involves intended and actuated behaviors to accommodate changes in environmental conditions, requiring the integration of intelligent sensing and control software. 
An example is the algorithm proposed in \cite{gabellieri2020grasp}, which determines the posture of a soft robotic hand to grasp objects with varying shapes and weights. 
Of note, active adaptability is not exclusive to soft robots and can also be implemented in rigid-link robots. 
However, the unique characteristics of soft robots, including their compact structures, surface deformation, and internal nonlinearities, introduce greater challenges for sensing and control. 
As a result, achieving active adaptability in soft robots is often more demanding than in rigid-link systems.}

\red{Internal adaptability includes the robustness and transferability of software or algorithms to accommodate variations in soft robot hardware.
Variability in hardware properties, resulting from material aging and manufacturing tolerances, represents an inherent limitation of soft robotic systems. When a robot system ages, continuing to use the same control algorithm may lead to degraded performance. Similarly, transferring the algorithm to another soft robot, even one with an identical design and fabricated using the same method, can result in reduced control accuracy due to manufacturing tolerances.  Intelligent algorithms can, to some extent, mitigate these issues, such as the controllers that compensate for differences in stiffness among robots produced under the same conditions \cite{chen2023hybrid} and an adaptive sensor calibration strategy for mass production \cite{kim2020adaptive}.
Beyond the robustness, some algorithms can be transferred across different hardware, such as controller transferable to soft robots with different module numbers \cite{chen2024novel} and different stiffness levels \cite{chen2025generalized}.}

\begin{figure}[!ht]
\centering
\includegraphics[width=\linewidth]{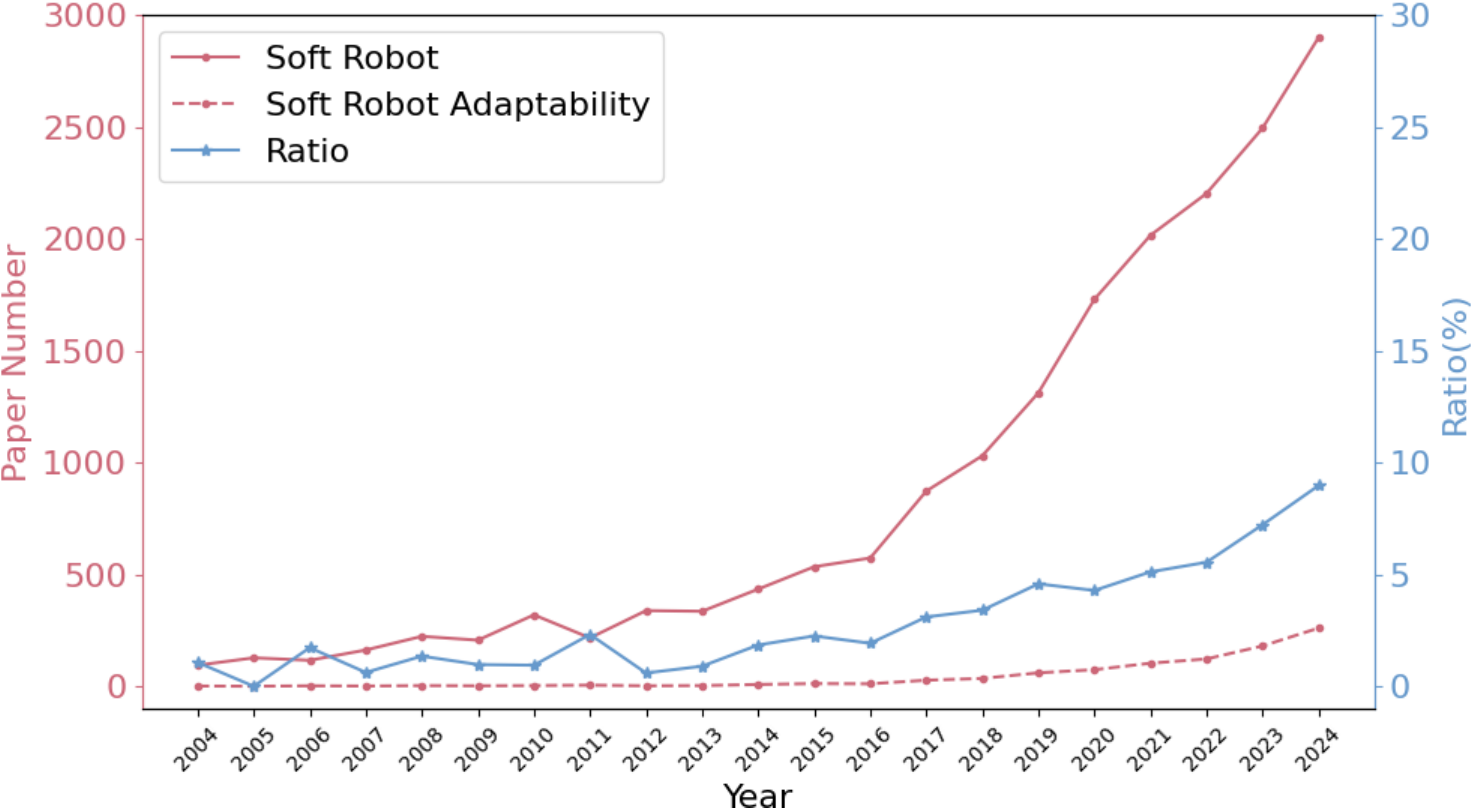}
\caption{\red{The number of publications containing the keywords ``Soft Robot" (red) and ``Soft Robot Adaptability" (dashed red), along with their ratio (blue). The data is based on search results from the Scopus database, limited to article title, abstract, and keywords. The figure illustrates the growing research interest in soft robotics, along with the increasing emphasis on adaptability as a key characteristic of soft robots.}}
\label{fig2}
\end{figure}

Compared to other surveys that cover the general aspects of soft robotics \cite{rus2015design, lee2017soft, chen2024data}, this review specifically highlights one of the most notable features of soft robots: adaptability. 
Using the Scopus database, we conducted an analysis of the number of publications related to soft robotics to reflect the field's research activity. 
The results, illustrated in Fig. \ref{fig2}, show that soft robotics research has grown significantly, as evidenced by a fivefold increase in publication numbers over the last ten years. 
Concurrently, as the field has developed, adaptability has gained increasing importance, as shown by the rising ratio of publications on soft robot adaptability compared to the total number of publications on soft robotics. 
Therefore, a review centered on adaptability is essential to summarize past research and provide insights into the future trajectory of soft robotics with this key feature in mind.
                                                    
The remainder of the paper is organized as follows: 
Section \ref{sec2} examines research dedicated to enhancing adaptability through advancements in robot design. 
Section \ref{sec3} reviews progress in soft sensors and the development of corresponding algorithms aimed at improving adaptability. 
Section \ref{sec4} provides an analysis of controllers that contribute to soft robot adaptability, encompassing both model-based and model-free approaches. 
Section \ref{sec5} explores various applications that leverage and benefit from the adaptability of soft robots. 
Section \ref{sec6} concludes with a comprehensive discussion involving the limitations of this area and forecasting potential future directions for soft robotics. 
Finally, Section \ref{sec7} summarizes this review.

\section{Robot Design for Adaptability}
\label{sec2}

\begin{figure*}[htbp]
  \centering
  \includegraphics[width=450pt]{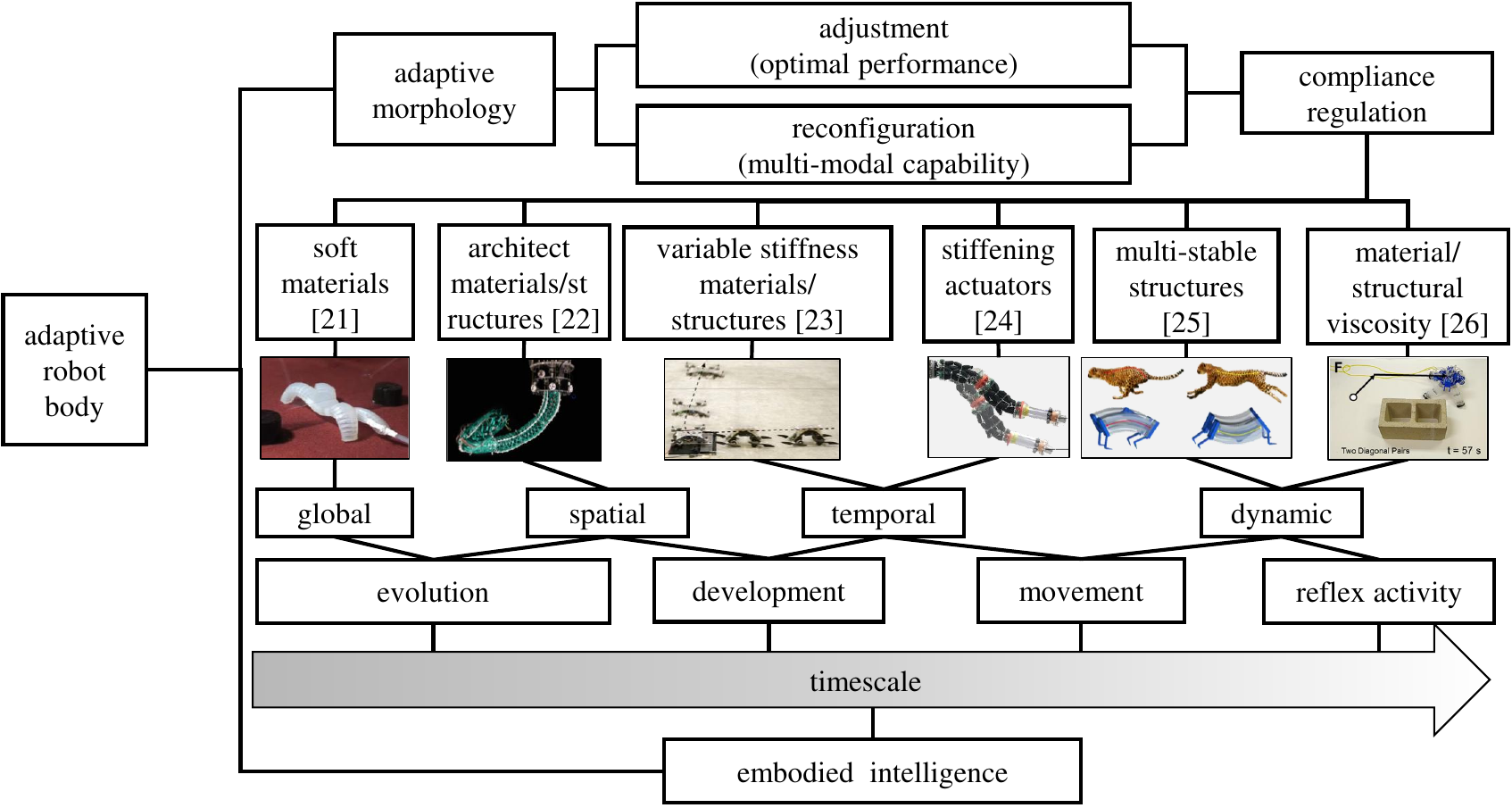}
  \caption{\red{Robot adaptability based on the body: Compliance regulation and Embodied Intelligence. Global soft crawling robot ~\cite{shepherd2011multigait}, Soft arm with varied spatial compliance~\cite{guan2023trimmed}, variable stiffness base on phase change materials~\cite{hwang2022shape} and antagonistic actuation~\cite{bruder2023increasing}, fast crawling based on bistable spine~\cite{tang2020leveraging}, auto gait based on ring oscillators ~\cite{drotman2021electronics}; All figures used with permission.}}
  \label{fig:body adapt}
\end{figure*}

Recent advancements in materials science and robotics technology have driven rapid progress in the field of soft robot adaptability. 
Soft robot technologies, characterized by their intrinsic compliance and adaptability, are revolutionizing the design paradigm towards mechanical intelligence~\cite{mintchev2016adaptive}. 
\red{The robot design contributed to soft robot adaptability is illustrated in Fig. \ref{fig:body adapt}.}

\subsection{Robotic adaptability based on compliance regulation}
Owing to the intrinsic compliance of soft materials and structures~\cite{kim2013soft}, many soft robots adapt to diverse environments and tasks through elastic deformation, adaptive configurations~\cite{rus2015design}, or multi-modal functions~\cite{hu2018small}. 
Meanwhile, rigid adaptive robots traditionally achieve adaptability through reconfigurable modular designs~\cite{zhao2024snail}.
Recent advancements integrate compliant mechanisms, such as elastic joints or deformable components, to reduce reliance on rigid parts and simplify actuation~\cite{onal2012modular}. 
Hybrid approaches, combining soft robotic principles with rigid architectures, enhance external adaptability while retaining precision~\cite{manti2016stiffening}.

Shape transformation is critical for adaptive morphology in both soft and rigid robots. 
While soft robots leverage intrinsic compliance to adjust their shape, this flexibility often trades off against load capacity~\cite{bruder2023increasing}. 
To address this, compliance customization strategies, such as spanning material, structural, and hybrid designs, enable tailored stiffness profiles for specific tasks~\cite{manti2016stiffening}.
For example, compliant mechanisms in rigid systems replace complex joints with elastic elements, enabling shape-morphing without sacrificing robustness~\cite{shan2023variable}. 
Similarly, soft robots incorporate variable stiffness materials (e.g., jamming, phase-change) to balance external adaptability and force transmission~\cite{dou2021soft}. 
These strategies highlight complementary pathways toward adaptable robotics, where soft and rigid paradigms converge through compliance regulation principles in different dimensions. 
The compliance regulation strategies applied for adaptability in soft robotics are detailed as follows:
   
{\subsubsection{Spatial Compliance Regulation -- Engineering Direction-Dependent Adaptability}
Spatial compliance regulation harnesses architected materials and structures to engineer robots with geometrically programmed stiffness, enabling external adaptability through directional mechanical heterogeneity. 
Tensegrity systems, such as those proposed by Zappetti et al.~\cite{zappetti2020phase}, integrate tensile and compressive elements to distribute compliance across predefined axes, while origami-inspired designs~\cite{vander2014origamibot} leverage foldable creases to achieve spatially varying deformation modes. 
Mechanical metamaterials further extend this paradigm, embedding anisotropic compliance into lattices or gradient architectures~\cite{grossi2021metarpillar}. 
For example, auxetic lattices with negative Poisson’s ratios simplify control by passively guiding deformation under load~\cite{mark2016auxetic}, whereas helicoidal structures optimize physical performance by balancing compliance and rigidity~\cite{guan2023trimmed}. 
By tailoring geometric stiffness, robots dynamically adapt to diverse tasks and unstructured environments, demonstrating how spatial compliance regulation bridges passive mechanics and active functionality and finally achieves external adaptability.}

\subsubsection{Temporal Compliance Regulation -- Real-Time Stiffness Reconfiguration}

Building on spatial design, temporal compliance regulation introduces time-dependent stiffness modulation, allowing robots to reconfigure mechanical properties in response to diverse real-time demands and achieve external adaptability to different tasks.
Phase-change materials (PCMs) exemplify this capability: low-melting-point alloys~\cite{shintake2015variable} and shape memory polymers (SMPs)~\cite{mohammadi2024sustainable} transition between rigid and compliant states via thermal stimuli, while shape memory alloys (SMAs) enable reversible stiffness through Joule heating~\cite{jiang2020variable}. 
Complementary to PCMs, jamming systems achieve rapid rigidity shifts by vacuum-driven particle immobilization~\cite{cheng2012design}. 

Critically, these materials are often integrated with actuation mechanisms, creating unified systems where motion and stiffness are co-regulated. 
Hydrostatic actuators, for instance, exploit fluid pressure to simultaneously drive shape morphing and modulate stiffness~\cite{giannaccini2014variable}, while antagonistic pneumatic chambers balance compliance through opposing forces~\cite{chen2020novel}. 
Such integration is exemplified by reconfigurable grippers that conform to various objects in a compliant state and rigidify for secure manipulation~\cite{linghu2020universal}, or robotic arms that switch between high payload capacity and confined-space navigation~\cite{gu2023self}. 
These advancements highlight how temporal regulation leverages material innovation for external adaptability, enabling task versatility in dynamic environments.

\subsubsection{\red{Dynamic Compliance Regulation -- Programming Transient Behaviors via Energy Dynamics} }
\red{Extending beyond static or time-varying stiffness, dynamic compliance regulation focuses on energy-mediated transient behaviors, where robotic motion is governed by energy storage and release, producing robust periodic motions passively adaptive to various environments.
Multi-stable mechanisms epitomize this approach: elastic snap-through instabilities, as demonstrated by Tang et al.~\cite{tang2020leveraging}, enable millisecond-scale locomotion on different surfaces by repetitively releasing stored strain energy.
Furthermore, oscillatory systems expand this paradigm with solar- or thermal-driven oscillators sustaining autonomous rhythmic motion~\cite{zhao2023sunlight}. 
These strategies embody principles of mechanical intelligence, where energy dynamics encode control logic directly into the robot’s structure and lead to designed responses to different interactions. 
For instance, electronics-free locomotion robots in \cite{drotman2021electronics} can avoid various obstacles by leveraging ring oscillators and passively adapting to external interactions without electronics and controllers.
By exploiting periodic dynamics, dynamic compliance regulation transcends traditional control paradigms, generating responses to various interactions with passive adaptability.}


\subsection{Synergistic Compliance regulation for Multifunctional Adaptability}
In the future, the integration of spatial, temporal, and dynamic compliance regulation may bridge passive mechanical responses and active behavioral control, enabling robots to achieve multifunctional external adaptability. 
For instance, a robotic arm might utilize spatially graded jamming structures to passively conform to cluttered environments through inherent compliance while temporally modulating stiffness to actively enhance payload capacity during object manipulation.
Concurrently, snap-through dynamics could enable rapid repositioning by harnessing stored elastic energy, blending passive instability with triggered actuation. 
Such hybrid systems will epitomize the convergence of geometric design, material innovation, and energy-aware control, where passive compliance ensures safe interaction and environmental adaptation while active regulation enables task-specific performance. 
Future advancements can focus on unifying these paradigms, like combining embodied intelligence with active sensing and control, to create robots that autonomously transition between adaptive and deliberate behaviors as responses to dynamic and unstructured settings.

This adaptability is hierarchically structured across timescales. 
In the future, evolutionary adaptation, though currently limited to simulation or prototyping iteration, may leverage spatial compliance principles such as modular robots with reconfigurable stiffness distribution to bridge simulated design with real-world applications, expanding the soft robotics paradigm. 
Developmental refinement may employ temporal regulation strategies like jamming layers or shape-memory polymers to optimize recurring tasks, such as grippers dynamically tuning stiffness to handle variable objects~\cite{gu2023self}. 
At the shortest timescale, reflexive action exploits dynamic mechanisms (e.g., snap-through bistability) to achieve rapid, energy-efficient responses, such as collision avoidance via elastic energy release~\cite{rothemund2018soft}. 
Together, these possible strategies can form a hierarchical framework: evolutionary designs establish adaptable morphologies, developmental tuning refines sustained functionality, and dynamic regulation enables self-regulated reflexes. 
By negotiating compliance across spatial, temporal, and energetic domains, this framework will emulate biological adaptability, reducing reliance on centralized control and advancing robots toward embodied intelligence \cite{hughes2022embodied}, where physical and computational adaptability converge to enable seamless interaction in unstructured environments.

\section{Sensing for Adaptability}
\label{sec3}

\begin{figure}[!ht]
\centering
\includegraphics[width=\linewidth]{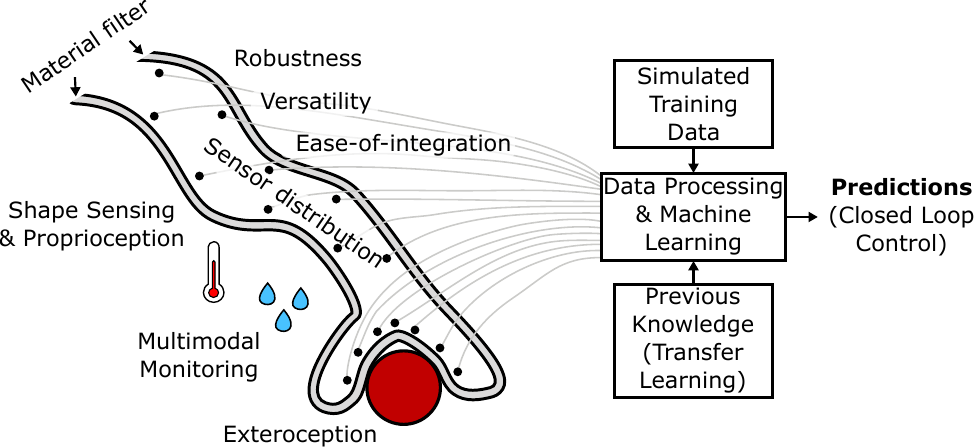}
\caption{\red{Key areas for adaptability in {the sensory system of} a soft robotic manipulator.}}
\label{fig:sensorfig}
\end{figure}

The appropriate design of a soft robot's sensory system is critical for its deployment and adaptation to a true range of environments, especially those for which it has no prior information. In these cases, extensive information must be gathered about the surroundings and the robot itself, relying on multiple sensing capabilities: robot tip localization \cite{ha2021robust}, pressure \& shape sensing of deformations caused by contact with new objects \cite{ward2018tactip}
, monitoring of environmental conditions \cite{HardmanDavid2022Sigh}, and high-speed reflexive responses \cite{partridge2020reflex}. {As environmental diversity increases during the exploration and active adaptability of soft robots, their need for sensing will be a critical bottleneck. Without this, functionalities will remain limited. However,} implementing sensing with minimal changes to a robot's design generates challenges in hardware and software design domains \cite{hegde_sensing_2023}, both of which are explored in this section.
\red{The diagram about sensing for adaptability is illustrated in Fig. \ref{fig:sensorfig}.}

\subsection{Sensor Hardware}
At the hardware level, longevity and compliance are essential. 
Soft sensors must integrate seamlessly into a soft robot's body stably without hindering its inherited morphologies. 
These robots undergo repeated cycles of stretching, bending, and flexing, which necessitate sensor longevity and flexibility. 
Their compliance generates external adaptability, ensuring that the material successfully conforms to unknown objects during interactions with new environments. 
They can be integrated as channels, applied as surface skins, or even serve as the robot's primary structural material to meet the requirements of different tasks.
The requirements of longevity and compliance have driven the ongoing development of highly compliant sensor materials. 
As reviewed by Hegde et al. \cite{hegde_sensing_2023}, these materials include liquid metals and stretchable waveguides, which can be embedded into an existing morphology. 
Additionally, conductive elastomer composites and hydrogels offer versatility. 

\red{In addition to sensing components in a robotic platform, soft sensors and skins can serve as an independent sensing system to achieve external adaptability.
They can be easily applied to different shapes and are likely to see more widespread use. 
This straightforward application allows them to be implemented across various objects and tasks without requiring fundamental changes to their underlying technologies. 
For example, Ohmura et al.'s sensing elements \cite{ohmura2006conformable} could be flexibly wrapped to curved surfaces while using a serial bus to reduce electrical connections. 
The ROBOSKIN project's modular elements \cite{billard2013roboskin} provided scalability to various large-area designs. 
Since their free surface areas and single-layer implementations promise easy applications to complex shapes, skins based on Electrical Impedance Tomography (EIT) are becoming increasingly popular \cite{LiuKai2020ASSf}, feasible for various shapes and interactions.}

Much as in the design of robot body mechanics, embodied intelligence can be applied and exploited during the design of soft sensor morphologies to minimize the computational processing required by external adaptability across various tasks. 
Sensor receptors in the human body are distributed to best provide useful information over a huge variety of tasks \cite{IidaFumiya2016Aosm}. 
A number of soft sensor works have taken biological inspiration in the design of their morphologies. 
Among these, An et al. use triboelectric whisker-inspired mechanoreceptors to detect 1.1 $\mu$N tactile forces in real time \cite{AnJie2021Whiskers}. 
Thuruthel et al. improve a fiber network's damage resilience through a bio-inspired optimization of its grid-based morphology \cite{thuruthel_joint_2020}. 
Appropriate design of the mechanical interface between sensor and environment can also lessen necessary computations: raw sensor signals can be filtered by a robot's soft body to best match their surroundings. 
For example, Hughes et al. use a jamming filter between the sensor and environment to adapt to different classification tasks \cite{hughes_online_2021}, while Costi et al. use a magneto-active elastomer filter to tune a sensor's mechanical stiffness in real-time \cite{costi_magneto-active_2022}. 
Though such filtering approaches aid both internal and external adaptabilities by simplifying the signals generated by a soft sensor, the filtered information must still be processed to produce a meaningful result. 
Therefore, sensor software is also a necessary component for adaptive designs.

\subsection{Sensor Software}
When collecting information for the external adaptability to new environments through physical interactions, shape sensing and distributed tactile sensing can be exploited to map a soft robot's surroundings \cite{Albini2021Exploiting}, facilitating closed-loop control. 
However, shape sensing presents significant computational difficulties: many soft sensors are already embedded within or adhered to complex 3D shapes and exhibit response nonlinearities, drift, and viscoelasticities. 
{Indeed, the tendency of many soft sensors to drift, degrade, and change material properties over time severely impedes their long-term robustness if continuous recalibrations are not carefully designed; existing papers consider signal robustness over hours or days rather than months or years.} 
These challenges must be accounted for when the signals are used for the reconstruction and decoupling of different stimuli. 
Machine learning techniques are typically employed to aid processing ~\cite{shih2020electronic, KimDaekyum2021Roml}
, enabling {robust} empirical modeling of the sensor state through ground truth training data. 
Still, if care is not explicitly taken during the design of these architectures and training methods, data-driven solutions may not be particularly {adaptive}: any changes in condition, task, or morphology could require total retraining from scratch rather than building upon pre-learned information. 

{To avoid this, transfer learning of pre-trained solutions can be employed to impart prior knowledge onto a new system for internal adaptability. 
Kim et al. introduce such a method for adaptive calibration in the face of long-term sensor usage, highlighting the approach's potential to address manufacturing tolerances during the larger-scale manufacture of soft robotic sensors \cite{kim2020adaptive}. 
Even without transfer learning, the continued application of recalibration over a sensor's lifetime can help to enhance robustness and deal with drift \cite{feng201wearable}. 
Rong et al. use deep transfer learning to transfer knowledge of a wearable device's response characteristics between gesture classification tasks, increasing learning efficiency by re-deploying feature extraction layers \cite{RONG2023114693}. 
Terryn et al. employ transfer learning to re-learn tasks after damage-heal cycles of self-healing soft sensors, ensuring {adaptability} on both the hardware and software levels \cite{TerrynLearning}. 
Here, the training data was physically collected; for the generation of more robust and transferable models, simulations can help to generate larger amounts of diverse data without facing the bottleneck of physical experimentation \cite{park2022biomimetic}.}

\section{Control Algorithms for Adaptability}
\label{sec4}

The inherent compliance and mechanism design of soft robots offer significant advantages in navigating complex environments and manipulating irregular objects. 
Meanwhile, the hardware flexibility often needs to be complemented by advanced control algorithms to fully realize their external adaptability. 
These algorithms enable soft robots to adapt their shape, movement, and interaction with the environments and objects.
Recently, some control algorithms have been proposed to cope with the variance in hardware characteristics across different platforms.
In this section, we explore the role of control algorithms in achieving internal and external adaptability. 
We summarize the adaptability through control approaches in Table \ref{table_ctrl}.

\subsection{Model-based Control}
Model-based control strategies apply physical models involving nonlinearity and hysteresis to represent the soft robot motion and achieve interaction with external environments.
These environments present additional complexities, including dynamic disturbances \cite{zhang2018vision} and intricate contour constraints \cite{della2020model}. 
To achieve external adaptability, model-based control strategies construct the environments with physical interaction models.
Recently, some works aim to achieve internal adaptability by updating the physical models.

\red{Model-based control approaches leverage various physical models like the Constant Curvature (CC) model \cite{della2020model} and Finite Element Method (FEM) \cite{zhang2018vision}. 
To enhance the active external adaptability of soft robots, model-based approaches rely on building physical models for the external environments and objects. 
For instance, a spring and damper model is utilized in \cite{della2020model} for contact modeling, and a Cartesian impedance controller is proposed based on this contact model. 
Equipped with such a model-based controller, a soft robot can safely trace along various surfaces. 
To grasp objects with different shapes, target objects are discretized into several boxes in \cite{gabellieri2020grasp}.
The grasping strategy selects the feasible grasp position based on the box model and controls the soft hand to grasp.}

In addition to modeling external environments, model-based controllers also show internal adaptability by updating the ideal models to match the real robots. 
One possible solution is parameter updating. 
The model parameters in Model Predictive Control (MPC) update according to the real robot feedback in \cite{tang2019novel}. 
The pseudo rigid robot model is applied in \cite{trumic2021adaptive} to match the piecewise constant curvature of soft robots with different stiffness levels through online parameter update.
Also, compensation outside the model-based controller can be applied for model matching. 
Overall, for both external and internal adaptability, model-based strategies depend on predefined models, and their performance will be highly affected by the accuracy of the physical models. However, there exists a fundamental trade-off: complex models may reflect real interactions but lack the flexibility for real-time updates, whereas concise models can update online to mitigate model mismatches but may sacrifice accuracy.

\subsection{Model-free Control}
In addition to the model-based methods mentioned above, many researchers have employed model-free approaches in soft or continuum robotics thanks to their ability to deal with highly non-linear and dynamic behaviors \cite{wu2024review, wu2021hysteresis}. 
Lightweight model-free controllers can update online and fit the mismatch caused by external environments and hardware variance.
Meanwhile, LfD and RL serve as effective learning strategies to cope with various and dynamic environments.

\red{The Jacobian controller is a model-free soft robot controller. Nonlinearity and hysteresis are addressed by the high-frequency online updates of the linear Jacobian matrix. 
In \cite{yip2014model}, the soft robot can track the trajectory even if affected by different unknown obstacles. 
A generalized Jacobian controller in \cite{chen2025generalized} can transfer across robots with different stiffness only relying on online parameter updating. 
In addition to the Jacobian-based controllers, some online updating statistical controllers also endow soft robots with internal and external adaptability. 
For instance, the online Gaussian Process Regression (GPR) controller in \cite{tang2020probabilistic} is validated in simulation, various real robots, and even under collision with obstacles.
GPR controllers in \cite{fang2019vision} can map between the robot state and actuation spaces online, and such a controller can achieve trajectory following tasks even under unknown and abrupt loads due to its external adaptability.}


LfD or imitation learning can be one of the candidates for enhancing the external adaptability of soft robots by enabling them to learn and replicate expert behaviors in dynamic conditions. 
When learning from expert demonstrations, soft robots can acquire flexible and compliant behaviors by observing how humans naturally adjust their movements in response to different conditions and tasks. 
A study demonstrated that, after receiving demonstrations from multiple humans using their own hands, the RBO Hand 2 exhibited strong performance across various manipulation tasks \cite{gupta2016learning}. 
LfD can also help soft robots improve their external adaptability in motion planning for navigating complex environments. 
For instance, one LfD method was proposed for the motion planning of multi-segment flexible and soft robotic manipulators, enabling it to automatically navigate through narrow spaces \cite{wang2016motion}. 

RL is a type of machine learning where an agent learns to make decisions by interacting with its environment and receiving feedback in the form of rewards or penalties. 
Through trial and error, the agent optimizes its actions to maximize long-term rewards, allowing it to adapt to dynamic and unpredictable conditions. 
This makes RL particularly useful for soft robots, which operate in environments where precise modeling is difficult and flexibility is essential. 
An RL-based adaptive control framework was proposed in \cite{li2022towards}, enabling precise control of a fluid-driven soft robot in continuous task space. 
The adaptability is first guaranteed by domain randomization in simulation and then enhanced by incremental learning in reality.
Finally, the internal adaptability is demonstrated through Sim2Real transferring, and the external adaptability is demonstrated through unpredicted external load. A robust path planning framework, Curriculum Generative Adversarial Imitation Learning (C-GAIL), is proposed in \cite{li2024robust} to navigate catheters through tortuous and deformable vessels by accounting for interactions with vessel walls and deformation. The method outperforms state-of-the-art approaches in both in-silico and in-vitro experiments

\red{Overall, similar to the model-based controllers, sophisticated strategies like RL require high costs considering time and explored space, but they can be transferred across different objects, environments, and tasks.
Concise controllers like the Jacobian controller can update online but can only be applied to low-level control.
These trade-offs highlight that there is no universal control solution that simultaneously optimizes adaptability, accuracy, and efficiency. Instead, controllers should be selected based on the specific requirements of the task, with appropriate compromises among these performance metrics.
In addition, a promising direction lies in hybrid approaches.
By integrating hierarchical frameworks, such as applying model-free RL for adaptive planning and model-based PCC controller for fast response, soft robots may guarantee both adaptability and efficiency.}

\begin{table*}
\centering
\caption{\red{Adaptability through control methods}}
{\color{black}
\begin{tabular}{>{\hspace{0pt}}m{0.08\linewidth}>{\centering\hspace{0pt}}m{0.063\linewidth}>{\hspace{0pt}}m{0.14\linewidth}>{\hspace{0pt}}m{0.08\linewidth}>{\hspace{0pt}}m{0.15\linewidth}>{\hspace{0pt}}m{0.32\linewidth}}
\hline
Control type & \multicolumn{1}{>{\hspace{0pt}}m{0.063\linewidth}}{Reference} & Method & Adaptability & Application & Validation\\ 
\hline
\multirow{4}{0.098\linewidth}{\hspace{0pt}Model-based control} & \cite{della2020model} & Pseudo rigid robot & External & External environment interaction & Simulation and reality to trace on different and complex surfaces\\ \hhline{~-----}
& {\cellcolor[rgb]{0.753,0.753,0.753}}\cite{gabellieri2020grasp} & {\cellcolor[rgb]{0.753,0.753,0.753}}Minimum volume bounding box & {\cellcolor[rgb]{0.753,0.753,0.753}}External & {\cellcolor[rgb]{0.753,0.753,0.753}}Grasping and manipulation & {\cellcolor[rgb]{0.753,0.753,0.753}}Objects with different shapes and weights\\ 
\cline{2-6}
& \cite{tang2019novel} & Model Predictive Control & Internal & Rehabilitation glove & Inaccurate model initialization and updating to achieve accurate model\\ 
\hhline{~-----}
& {\cellcolor[rgb]{0.753,0.753,0.753}}\cite{trumic2021adaptive} & {\cellcolor[rgb]{0.753,0.753,0.753}}PCC & {\cellcolor[rgb]{0.753,0.753,0.753}}Internal & {\cellcolor[rgb]{0.753,0.753,0.753}}Segmented robot control & {\cellcolor[rgb]{0.753,0.753,0.753}}Robots with different stiffness\\ 

\hline

\multirow{6}{0.098\linewidth}{\hspace{0pt}Model-free control} & \cite{yip2014model} & Jacobian controller & External & Soft continuum robot control & Environments including single and multiple constraints\\ 
\cline{2-6}
& {\cellcolor[rgb]{0.753,0.753,0.753}}\cite{chen2025generalized} & {\cellcolor[rgb]{0.753,0.753,0.753}}Generalized Jacobian control & {\cellcolor[rgb]{0.753,0.753,0.753}}Internal & {\cellcolor[rgb]{0.753,0.753,0.753}}Soft continuum robot control & {\cellcolor[rgb]{0.753,0.753,0.753}}Robots with different stiffness and frequencies \\ 
\hhline{~-----}
& \cite{tang2020probabilistic} & Gaussian Process Regression & Internal & Soft finger control & Robots with different materials \\ 
\hhline{~-----}
& {\cellcolor[rgb]{0.753,0.753,0.753}}\cite{gupta2016learning} & {\cellcolor[rgb]{0.753,0.753,0.753}}Learning from Demonstration & {\cellcolor[rgb]{0.753,0.753,0.753}}External & {\cellcolor[rgb]{0.753,0.753,0.753}}Manipulation & {\cellcolor[rgb]{0.753,0.753,0.753}}Three manipulation tasks, i.e. turning a valve, manipulating an abacus, and grasping\\ 
\hhline{~-----}
& \cite{li2022towards} & Reinforcement learning & External & Soft continuum robot control & Path following under varying load\\ 
\hline
\end{tabular}}
\label{table_ctrl}
\end{table*}

\section{Soft Robots' Adaptability Across Different Applications}
\label{sec5}
\subsection{Interventions and Surgeries}
The adaptability of soft robots plays a crucial role in enhancing the safety and effectiveness of interventional and surgical procedures. 
In various surgical applications, the inherent flexibility of soft robots allows them to conform precisely to the complex and dynamic environments within the human body. 
This adaptability enables them to reach areas that are otherwise inaccessible with conventional, rigid instruments while interacting gently and accurately with surrounding tissues. 
This is essential in reducing the risk of complications, such as tissue damage or perforation, which could have severe consequences. 
Furthermore, since soft robots reduce concerns about applying excessive force to tissues, the mental load on surgeons is also lessened. 
\paragraph{Enhancing Surgical Safety} 
The most direct impact of the passive adaptability of soft robots on surgical and interventional procedures is the significant enhancement of safety during these processes. 
For instance, in intraluminal interventions, where precision and delicacy are paramount, the adaptability of soft robots, whether achieved through compliant control algorithms \cite{wu2022deep, jakes2019model}, the use of soft materials, or compliant mechanisms, proves to be exceptionally valuable. 
The lack of adaptability in surgical instruments can result in accidental perforations, a serious complication that may cause internal bleeding, infection \cite{johnson2019much}. 
These complications could lead to extended hospitalization and prolonged recovery times. 
In comparison, soft robots, with their built-in passive adaptability, are less prone to causing harmful movements during a failure, providing a safer response in such situations \cite{kwok2022soft}.

\paragraph{Improving Access to Challenging Anatomical Sites} 
The active adaptability of soft robots allows them to navigate complex anatomical pathways, providing surgeons with greater flexibility in reaching challenging sites \cite{zhang2020review}. 
Unlike rigid instruments, which are limited in their ability to move through curved or confined spaces, soft robots can bend and adjust to fit into narrow or winding areas, following the body’s natural contours with ease. 
This increases their reach and dexterity, enabling them to access deep or otherwise difficult-to-reach areas with minimal incisions, which is particularly advantageous in minimally invasive procedures. 
Another feature that highlights the active adaptability of soft robots is their ability to be squeezed.
For instance, the digestive tract often experiences luminal stenosis, making it difficult to introduce conventional endoscopes. 
A squeezable soft robot can effectively navigate through these constricted areas, allowing for both diagnostic and therapeutic interventions in such situations.


\paragraph{Reducing Clinician Mental Load} 
Traditional surgical procedures require surgeons to maintain a high level of concentration and manual dexterity, which can lead to cognitive and physical fatigue, especially during lengthy or complex procedures \cite{petrut2020mental}. 
The constant need to adjust and manipulate instruments to navigate the body's intricate structures further adds to this strain, increasing the potential for errors.
Soft robots, with their adaptive capabilities, help alleviate this burden by conforming naturally to the complex and dynamic environments within the human body. 
Their flexibility allows for smoother and more intuitive movements, meaning the surgeon spends less time and effort on manually adjusting the instruments. 
This passive adaptability is particularly valuable in intraluminal interventions, where clinicians typically spend significant time and effort steering the robot tip to prevent its acute angle from perforating the lumen wall. 


\subsection{Wearable Soft Robotics}
Wearable soft robots have been leveraged for various applications, such as rehabilitation, assistance,
and augmentation. 
Due to their compliance and external adaptability, soft robots can perfectly suit human body shape and deform adaptively to human motion. 
The robot design, sensing algorithms, and controllers all contribute to the adaptability of wearable soft robots. 
Several reviews have summarized the development of wearable soft robots\cite{xiloyannis2021soft}.

The hardware of exoskeleton has evolved from heavy and structured rigid robots to lightweight and adaptive wearable soft robots. 
Wearable soft robots are designed to be fixed on the key part of the human body instead of covering the whole body to adapt to different body sizes. 
For instance, the lower limb exoskeleton \cite{di2019design} is attached to the wrist, thigh, and lower leg to avoid joints. 
Fixed parts in soft gloves \cite{butzer2021fully} are connected by actively sliding springs to adapt the hand motions.

Based on the adaptive exoskeleton design, sensing also plays a significant role in adaptability. 
Conformal tactile textiles \cite{luo2021learning} are leveraged to make gloves, socks, and shirts adaptive to the different shapes of human body components. 
The wearable soft sensor feedback can detect various human interactions with external environments. 
Different types of sensors such as Electromyography (EMG) and Inertial Measurement Unit (IMU) \cite{li2022multi} can be integrated into exoskeletons for sensing fusion and adapting to diverse patients.

Considering controller for adaptability, Finite-State Machine (FSM) is widely applied in wearable soft robots. 
This strategy is leveraged in the glove for the assistance of different daily living activities \cite{polygerinos2015emg}. 
Machine learning methods like neural networks \cite{ha2018use} can be adaptive to different users without the requirement of accurate human models.

\subsection{Locomotion}
Locomotion in unstructured environments requires the external adaptability of mobile soft robots. 
Due to their flexibility and lightweight, mobile soft robots have shown excellent performance on the field, underwater, and on various surfaces compared to rigid robots. 
Crawling, flying, jumping, and swimming are the main locomotion approaches of mobile soft robots.
Some surveys summarize the developments of the mobile soft robots \cite{calisti2017fundamentals}.

The motions of crawling soft robots are inspired by animals like snakes \cite{qi2020novel} and starfish \cite{yang2021starfish}. 
Worm-inspired soft robots \cite{tang2018switchable} can adhere to grounds and walls with different materials thanks to the soft adhesion actuators adaptive to different surfaces. 
The starfish robot in \cite{yang2021starfish} can perform omnidirectional movements on various harsh environments like wet and rough surfaces. 
Exploiting the traveling-wave motion, the snake robot in \cite{qi2020novel} can move through pipelines with turnings. 
The above robots have demonstrated their external adaptability through their ability to navigate diverse environments.

Similarly, soft swimming robots are bioinspired and developed by the animal motion patterns, such as fish \cite{katzschmann2018exploration} and jellyfish \cite{godaba2016soft}. 
The fish robot in \cite{katzschmann2018exploration} has demonstrated its external adaptability in complex and unstructured oceans and seabeds. 
Light soft jellyfish robot in \cite{godaba2016soft} can perform both vertical and horizontal motions underwater. 
In addition, a nonbiomorphic swimming robot in \cite{liu2023nonbiomorphic} can perform multiple maneuvers and hence is able to propel in cluttered underwater environments, including obstacles and tunnels. 

Moreover, there are locomotion robots composed of some soft components, which enhance their external adaptability. 
A compact jumping robot \cite{wang2023amphibious} can jump to cross a ring and over an obstacle through different targeted motions.
Soft pneumatic bars are equipped on an aerial robot \cite{nguyen2023soft}, which can be resilient to external collision and impact. 
Universal granular grippers are installed on a legged robot \cite{hauser2018compliant} as feet to be adaptive to different surfaces.

\subsection{Manipulation}

As one of the most significant soft robot applications in the industry, soft grippers have shown unique advantages over conventional rigid grippers. 
Grasping and manipulating objects with different morphologies and properties require the gripper to possess a high degree of passive external adaptability. 
Thanks to compliance and passive deformation, soft robot grippers can perform stable and safe grasps on objects with different shapes, weights, and stiffness.
There are several surveys focusing on soft robot grippers \cite{hughes2016soft}.

The external adaptability to different objects is mainly achieved by the passive deformation of soft grippers.
Silicone fingers in \cite{zhou2017soft} show different passive deformation patterns on different object shapes. 
Furthermore, the gripper can perform stable grasps thanks to the surface patterned feature. 
Inspired by human hand grasp synergies, the Pisa/IIT SoftHand 2 is proposed in \cite{della2018toward}. 
Different grasp postures can be performed with the combination of only two main natural postural synergies, thanks to the passive deformation caused by the interaction. 
In addition to passive deformation, stiffness change endows granular jamming grippers with shape adaptability. 
The granule-filled bag in \cite{brown2010universal} first adapts to the object due to the softness, then constrains the object motion and grasp by evacuating.

In addition to shape compliance, stiffness adaptability is motivated by the different grasping requirements. 
Some objects are fragile and require gentle grasps like tofu \cite{zhou2017soft}. 
Meanwhile, some heavy objects require stable and strong grasps. 
The rigid-soft gripper in \cite{zhou2017soft} produces gentle grasps leveraging the soft pads and guarantees grasp stability leveraging the rigid clamps. 
Similarly, the soft gripper in \cite{subramaniam2020design} composes soft skins and rigid bones, which can grasp an egg safely and a heavy bottle stably. 
In addition to these passive adaptability works, the gripper in \cite{li2017variable} can actively adjust its stiffness and manage proper grasps.

\section{Limitation and Future Work}
\label{sec6}
While this review presents various adaptability technologies in soft robots and highlights their applications across diverse fields, significant challenges persist in further advancement of adaptability. 
In this section, we discuss the limitations and potential future directions of these technologies.

\subsection{Limitations}
\red{While soft robots offer significant advantages, they also possess inherent limitations and are not a one-size-fits-all solution for every robotic application.
Compared to rigid robots, soft robotics face limitations such as low payload due to their compliant nature and inconsistent accuracy resulting from the high manufacturing tolerance.
Although adaptive technologies may alleviate some issues, such as calibration for sensor variance \cite{kim2020adaptive}, certain limitations remain intrinsic to soft robotics.
In the future development of soft robotics, instead of restricting the scope and focus solely to soft materials and flexible structures, robotic systems that selectively integrate both rigid and soft components based on application needs may better harness the advantages of each, potentially paving the way for a new subfield within soft robotics.}

\red{Furthermore, while adaptability is a crucial feature in soft robotics, it is not the sole consideration.
A careful balance among various properties is necessary during both hardware manufacturing and software development. 
For instance, soft tactile sensors are typically designed as fingertip sensors to enable contact with diverse surfaces \cite{zhang2022hardware}, whereas a roller sensor is designed specifically for plane sensing \cite{cao2023touchroller}. 
Although it can only be applied to planes and loses adaptability, the targeted design enhances the sensing areas and speeds.
Overall, adaptive soft robots are not universal solutions. 
Adaptability, along with other factors, should be carefully weighed according to the task requirements during the robot platform design.}

\subsection{Future Work}
\red{External adaptability, particularly active external adaptability, serves as one of the most valuable characteristics in soft robotics, which can be demonstrated from the applications in Section \ref{sec5}.
Novel materials, robot structures, and the corresponding algorithms may enhance the external ability, such as multifunctional and independent modules. 
For instance, modular soft robot arms have exhibited adaptability in navigating clutter environments compared to single module robots \cite{marchese2014whole}.
However, current modular designs are often constrained by centralized actuation sources and actuation lines such as cables or airlines, which limit the number of modules and the range of possible morphological structures.
Additionally, the modules are typically identical and designed for bending, restricting their versatility in performing diverse manipulation tasks.
Endowing modules with independent actuation sources and specialized functionalities would allow for a greater variety of configurations and broader application potential.
Based on the novel hardware platform, the corresponding algorithms, like the multi-agent strategy \cite{malley2020eciton}, are essential to enable assembly in diverse patterns and fully achieve external active adaptability.}

\red{Moreover, embodied intelligence offers a powerful strategy for analyzing and enhancing robotic adaptability, particularly in the face of diverse and evolving compliance regulations~\cite{hughes2022embodied}.  
By leveraging the concept of embodied intelligence, robots can exhibit more natural and responsive behaviors by integrating mechanical elements that mimic biological systems, such as adaptive compliance and the autonomic nervous system. Elements based on instability, viscosity, and oscillation can function similarly to the sympathetic and parasympathetic nervous systems, enabling rapid mobilization or slow dampening responses~\cite{porges1997emotion}. By combining embodied intelligence, soft robots can exhibit a wide range of adaptive behaviors, from long-term evolutionary adaptations to short-term reflexive reactions~\cite{iida2023timescales}.}

\red{Compared to external adaptability, internal adaptability is still in its nascency.
Instead of a benefit of soft robotics, robustness in internal adaptability serves as a compensatory solution to the inherent limitations such as aging and high manufacturing tolerances.
However, these limitations may be solved by alternative approaches, such as durable materials and standard manufacturing processes.
As soft robotics technology advances, the randomness of hardware properties will be alleviated, reducing the necessity for robustness in control algorithms.
Meanwhile, the increasing diversity of hardware platforms may elevate the importance of transferability.
In this context, online updating based on feedback will play a crucial role in enhancing transferability, and lightweight learning strategies may be introduced specifically for long-term and cross-platform usage. }

\section{Conclusion}
\label{sec7}
\red{
This survey explores one of the most critical features of soft robotics: adaptability. 
Based on the existing works related to adaptability, it can be categorized into external adaptability targeted at environments and objects and internal adaptability targeted at robotic system components. 
External adaptability is further divided into passive and active adaptability, while internal adaptability encompasses robust and transferable adaptability.
The increasing emphasis on adaptability in soft robotics research underscores the necessity of this review.
}

\red{
Subsequently, this review discusses how adaptability is achieved through robot design, sensing, and control. 
In soft robotics, design plays a foundational role in enabling external adaptability, particularly through passive adaptation. This is achieved by carefully engineering both the soft materials and compliant structures to provide spatial, temporal, and dynamic compliance.
Soft sensors are adaptive to different environments and objects due to their compliance, and sensing algorithms demonstrate their robustness to hardware primarily based on learning algorithms. 
Control strategies enhance external adaptability through external environment modeling and high-level learning strategy and internal adaptability through online updating mechanisms.
In addition, we list real-world applications utilizing soft robot adaptability, including surgery, wearable devices, locomotion, and manipulation.
We also discuss the limitations of soft robot adaptability and outline potential future directions in this field.
}

\section*{Acknowledgement}
This review is inspired by the workshop ``The SOFT Frontier: Adaptive Technologies in Soft Robotics" at the 2024 IEEE/RSJ International Conference on Intelligent Robots and Systems (IROS).

\bibliographystyle{IEEEtran}
\bibliography{IEEEabrv,references}
\vfill

\end{document}